# Pre-processing for Triangulation of Probabilistic Networks

**Hans L. Bodlaender,**
Institute of Information and Computing
Sciences, Utrecht University,
P.O. Box 80.089, 3508 TB Utrecht,
The Netherlands
hans@cs.uu.nl

**Arie M.C.A. Koster,**
Konrad-Zuse-Zentrum
für Informationstechnik
Berlin, Takustrasse 7,
D-14195 Berlin-Dahlem, Germany
koster@zib.de

**Frank van den Eijkhof,   Linda C. van der Gaag**
Institute of Information and Computing
Sciences, Utrecht University,
P.O. Box 80.089, 3508 TB Utrecht,
The Netherlands
{frankvde,linda}@cs.uu.nl

## Abstract

The currently most efficient algorithm for inference with a probabilistic network builds upon a triangulation of a network's graph. In this paper, we show that pre-processing can help in finding good triangulations for probabilistic networks, that is, triangulations with a minimal maximum clique size. We provide a set of rules for stepwise reducing a graph. The reduction allows us to solve the triangulation problem on a smaller graph. From the smaller graph's triangulation, a triangulation of the original graph is obtained by reversing the reduction steps. Our experimental results show that the graphs of some well-known real-life probabilistic networks can be triangulated optimally just by pre-processing; for other networks, huge reductions in size are obtained.

## 1 Introduction

The currently most efficient algorithm for inference with a probabilistic network is the *junction-tree propagation* algorithm that builds upon a *triangulation* of a network's moralised graph [1, 2]. The running time of this algorithm depends on the triangulation used. In general, it is hard to find a triangulation for which this running time is minimal. As there is a strong relationship between the running time of the algorithm and the maximum of the triangulation's clique sizes, for real-life networks triangulations are sought for which this maximum is minimal. The minimum of the maximum clique size over all triangulations of a graph is a well-studied notion, both by researchers in the field of probabilistic networks and by researchers in graph theory and graph algorithms. In the latter field of research, the notion of *treewidth* is used to denote this minimum minus one. Unfortunately, computing the treewidth of a given graph is an NP-complete problem [3].

When solving hard combinatorial problems, *pre-processing* is often profitable. The basic idea is to reduce the size of the problem under study, using relatively little computation time and without losing optimality. The smaller, and presumably easier, problem is subsequently solved. In this paper, we discuss pre-processing for triangulation of probabilistic networks. We provide a set of rules for stepwise reducing the problem of finding a triangulation for a network's moralised graph with minimal maximum clique size, to the same problem on a smaller graph. Various algorithms can then be used to solve the smaller problem. Given a triangulation of the smaller graph, a triangulation of the original graph is obtained by reversing the reduction steps. Our reduction rules are guaranteed not to destroy optimality with respect to maximum clique size. Experiments with pre-processing revealed that our rules can effectively reduce the problem size for various real-life probabilistic networks. In fact, the graphs of some well-known networks are triangulated optimally just by pre-processing.

In this paper, we do not address the second phase in the pre-processing approach outlined above, that is, we do not address actually constructing triangulations with a minimal or close to minimal maximum clique size. Recent research results indicate, however, that for small graphs optimal triangulations can be feasibly computed. Building upon a variant of an algorithm by Arnborg, Corneil, and Proskurowski [3], K. Shoikhet and D. Geiger performed various experiments on randomly generated graphs [4]. Their results indicate that this algorithm allows for computing optimal triangulations of graphs with up to 100 vertices.

The paper is organised as follows. In Section 2, we review some basic definitions. In Section 3, we present our pre-processing rules. The computational model in which these rules are employed, is discussed in Section 4. In Section 5, we report on our experiments with well-known real-life probabilistic networks. The paper ends with some conclusions and directions for further research in Section 6.

## 2 Definitions

The currently most efficient algorithm for probabilistic inference operates on a junction tree that is derived from a triangulation of the moralisation of the digraph of a probabilistic network. We review the basic definitions involved.

Let $G = (V, A)$ be a directed acyclic graph. The *moralisation* of $G$ is the undirected graph $M(G)$ obtained from $G$ by adding edges between every pair of non-adjacent vertices that have a common successor (vertices $v$ and $w$ have a common successor if there is a vertex $x$ with $(v, x) \in A$ and $(w, x) \in A$), and then dropping the arcs' directions.

Let $G = (V, E)$ be an undirected graph. A set of vertices $W \subseteq V$ is called a *clique* in $G$ if there is an edge between every pair of disjoint vertices from $W$; the cardinality of $W$ is the clique's *size*. For a set of vertices $W \subseteq V$, the *subgraph induced* by $W$ is the graph $G[W] = (W, (W \times W) \cap E)$; for a single vertex $v$, we write $G - v$ to denote $G[V - \{v\}]$. The graph $G$ is *triangulated* if it does not contain an induced subgraph that is a simple cycle of length at least four. A *triangulation* of $G$ is a triangulated graph $H(G)$ that contains $G$ as a subgraph. The *treewidth of the triangulation* $H(G)$ of $G$ is the maximum clique size in $H(G)$ minus 1. The *treewidth* of $G$, denoted $\tau(G)$, is the minimum treewidth over all triangulations of $G$.

A graph $H$ is a *minor* of a graph $G$ if $H$ can be obtained from $G$ by zero or more vertex deletions, edge deletions, and edge contractions (edge contraction is the operation that replaces two adjacent vertices $v$ and $w$ by a single vertex that is connected to all neighbours of $v$ and $w$). It is well known (see for example [5]), that the treewidth of a minor of $G$ is never larger than the treewidth of $G$ itself.

A *linear ordering* of an undirected graph $G = (V, E)$ is a bijection $V \leftrightarrow \{1, \ldots, |V|\}$. For $v \in V$ and a linear ordering $f$ of $G - v$, we denote by $(v; f)$ the linear ordering $f'$ of $G$ that is obtained by adding $v$ at the beginning of $f$, that is, $f'(v) = 1$ and, for all $w \neq v$, $f'(w) = f(w) + 1$. A linear ordering $f$ is a *perfect elimination scheme* for $G$ if, for each $v \in V$, its higher ordered neighbours form a clique, that is, if every pair of distinct vertices in the set $\{w \in V \mid \{v, w\} \in E \text{ and } f(v) < f(w)\}$ is adjacent. It is well known (see for example [6]), that a graph is triangulated if and only if it allows a perfect elimination scheme.

For a graph $G$ and a linear ordering $f$ of $G$, there is a unique minimal triangulation $H(G)$ of $G$ that has $f$ for its perfect elimination scheme. This triangulation, which we term the *fill-in given $f$*, can be constructed by, for $i = 1, \ldots, |V|$, turning the set of higher numbered neighbours of $f^{-1}(i)$ into a clique. The maximum clique size minus 1 of this fill-in is called the *treewidth of $f$*. The treewidth of a linear ordering of a triangulated graph equals the maximum number of higher numbered neighbours of a vertex [6].

To conclude, a *junction tree* of an undirected graph $G = (V, E)$ is a tree $T = (I, F)$, where every node $i \in I$ has associated a vertex set $V_i$, such that the following two properties hold: the set $\{V_i \mid i \in I\}$ equals the set of maximal cliques in $G$ and, for each vertex $v$, the set $T_v = \{i \mid v \in V_i\}$ constitutes a connected subtree of $T$. It is well known (see for example [6]), that a graph is triangulated if and only if it has a junction tree.

## 3 Safe reduction rules

Pre-processing a probabilistic network for triangulation builds upon a set of *reduction rules*. These rules allow for stepwise reducing a network's moralised graph to another graph with fewer vertices. The steps applied during the reduction can be reversed, thereby enabling us to compute a triangulation of the original graph from a triangulation of the smaller graph. In this section, we discuss the various rules; a discussion of the computational method in which these rules are employed, is deferred to Section 4.

During a graph's reduction, we maintain a stack of eliminated vertices and an integer *low* that gives a lower bound for the treewidth of the original graph. Application of a reduction rule serves to modify the current graph $G$ to $G'$ and to possibly update *low* to *low'*. We say that the rule is *safe* if $\max(\tau(G), low) = \max(\tau(G'), low')$. By applying safe rules, therefore, we have as an invariant that the treewidth of the original graph equals the maximum of the treewidth of the reduced graph and the value *low*. In the sequel, we assume that the original moralised graph has at least one edge and that *low* is initialised at 1.

Our first reduction rule applies to simplicial vertices. A vertex $v$ is *simplicial* in an undirected graph $G$ if the neighbours of $v$ form a clique in $G$.

**Lemma 1** *Let $G$ be an undirected graph and let $v$ be a simplicial vertex in $G$ with degree $d \geq 0$. Then,*

- $\tau(G) = \max(d, \tau(G - v))$;
- *there is a linear ordering $(v; f)$ of $G$ of minimum treewidth, where $f$ is a linear ordering of $G - v$ of treewidth at most $\max(d, \tau(G - v))$.*

**Proof.** Since $G$ contains a clique of size $d + 1$, we have that $\tau(G) \geq d$. We further observe that $\tau(G) \geq \tau(G - v)$, because $G - v$ is a minor of $G$. We therefore have that $\tau(G) \geq \max(d, \tau(G - v))$. Now, let $f$ be a linear ordering of $G - v$ of treewidth $k \leq \max(d, \tau(G - v))$. Let $H$ be the fill-in of $G - v$ given $f$. Adding vertex $v$ and its (formerly) incident edges to $H$ yields a graph $H'$ that is still triangulated: as every pair of neighbours of $v$ is adjacent, $v$ cannot belong to a simple (chordless) cycle of length at least four. The maximum clique size of $H'$ therefore equals the maximum of $d+1$ and $k+1$. Hence, $\tau(G) \leq \max(d, \tau(G - v))$, from which we conclude the first property stated in the lemma. To prove the second property, we observe that the linear ordering $(v; f)$ is a perfect elimination scheme for $H'$, as removal of $v$ upon computing the fill-in of $H'$ does not create any additional edges. □

Our first reduction rule, illustrated in Figure 1, now is:

**Reduction Rule 1: Simplicial vertex rule**
Let $v$ be a simplicial vertex of degree $d \geq 0$.
Remove $v$.
Set *low* to $\max(low, d)$.

From Lemma 1 we have that the *simplicial vertex rule* is safe. The second property stated in the lemma further provides for the rule's reversal when computing a triangulation of the original graph from one of the reduced graph.

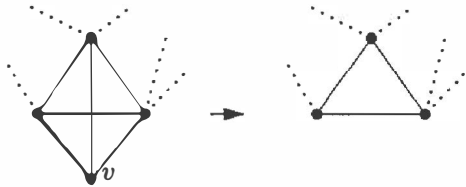

Figure 1: The *simplicial vertex rule*.

Because the digraph $G$ of a probabilistic network is moralised before it is triangulated, it is likely to give rise to many simplicial vertices. We consider a vertex $v$ with outdegree zero in $G$. Since all neighbours of $v$ have an arc pointing into $v$, moralisation will connect every two neighbours that are not yet adjacent, thereby effectively turning $v$ into a simplicial vertex. The *simplicial vertex rule* will thus remove at least all vertices that have outdegree zero in the network's original digraph. As every directed acyclic graph has at least one vertex of outdegree zero, at least one reduction will be performed. As the reduced graph need not be the moralisation of a directed acyclic graph, it is possible that no further reductions can be applied.

The digraph $G$ of a probabilistic network may also include vertices with indegree zero and outdegree one. These vertices will always be simplicial in the moralisation of $G$. We consider a vertex $v$ with indegree zero and a single arc pointing into a vertex $w$. In the moralisation of $G$, $w$ and its (former) predecessors constitute a clique. As all neighbours of $v$ belong to this clique, $v$ is simplicial.

A special case of the *simplicial vertex rule* applies to vertices of degree 1; it is termed the *twig rule*, after [7].

**Reduction Rule 1a: Twig rule**
Let $v$ be a vertex of degree 1.
Remove $v$.

The *twig rule* is based upon the observation that vertices of degree one are always simplicial. Another special case is the *islet* rule that serves to remove vertices of degree zero. We would like to note that many heuristic triangulation algorithms, such as the algorithm described in [8], remove simplicial vertices.

Our second reduction rule applies to almost simplicial vertices. A vertex $v$ is *almost simplicial* in an undirected graph $G$ if there is a neighbour $w$ of $v$ such that all other neighbours of $v$ form a clique in $G$. Figure 2 illustrates the basic idea. As we allow other neighbours of $v$ to be adjacent to $w$, simplicial vertices are also almost simplicial.

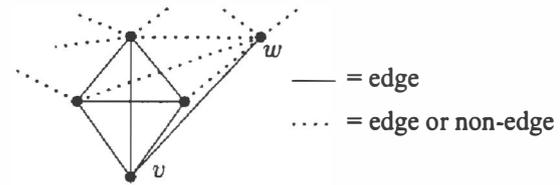

Figure 2: An almost simplicial vertex.

**Lemma 2** *Let $G$ be an undirected graph and let $v$ be an almost simplicial vertex in $G$ with degree $d \geq 0$. Let $G'$ be the graph that is obtained from $G$ by turning the neighbours of $v$ into a clique and then removing $v$. Then,*

- $\tau(G') \leq \tau(G)$ *and* $\tau(G) \leq \max(d, \tau(G'))$;
- *the linear ordering $(v; f)$ of $G$, with $f$ a linear ordering of $G'$ of treewidth at most $\max(d, \tau(G'))$, has treewidth at most $\max(d, \tau(G'))$.*

**Proof.** Let $w$ be a neighbour of $v$ such that the other neighbours of $v$ form a clique. As we can obtain $G'$ from $G$ by contracting the edge $\{v, w\}$, $G'$ is a minor of $G$. We therefore have that $\tau(G') \leq \tau(G)$. Now, let $f$ be a linear ordering of $G'$ of treewidth $k \leq \max(d, \tau(G'))$. Let $H$ be the fill-in of $G'$ given $f$. If we add $v$ and its (formerly) adjacent edges to $H$, then $v$ is simplicial in the resulting graph $H'$. Using Lemma 1, we find that $\tau(H') = \max(k, d)$. The two properties stated in the lemma now follow. □

Our second reduction rule, illustrated in Figure 3, is:

**Reduction Rule 2: Almost simplicial vertex rule**
Let $v$ be an almost simplicial vertex of degree $d \geq 0$.
If $low \geq d$, then
    add an edge between every pair of non-adjacent neighbours of $v$;
    remove $v$.

Building upon Lemma 2 we find that the *almost simplicial vertex rule* is safe. Suppose that we have $G$ and *low* before, and $G'$ and *low'* after application of the rule. Then, $\tau(G') \leq \tau(G)$, $\tau(G) \leq \max(d, \tau(G'))$, and $d \leq low = low'$. We conclude that $\max(\tau(G), low) = \max(\tau(G'), low')$. Examples can be constructed, unfortunately, that show that the rule is not safe for $low < d$.

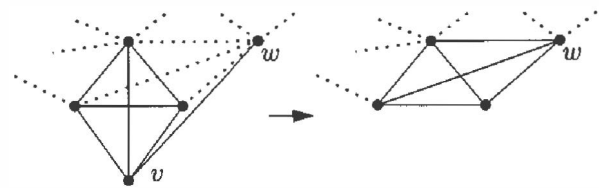

Figure 3: The *almost simplicial vertex rule*.

A special case of the *almost simplicial vertex rule* applies to vertices of degree two. A vertex of degree two is, by definition, almost simplicial and we can therefore replace it by an edge between its neighbours, provided that the original graph has treewidth at least two. The resulting rule, illustrated in Figure 4, is called the *series rule*, after [7].

**Reduction Rule 2a: Series rule**
Let $v$ be a vertex of degree 2.
If $low \geq 2$, then
　add an edge between the neighbours of $v$, if
　they are not already adjacent;
　remove $v$.

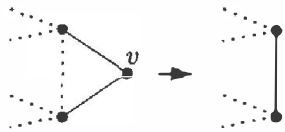

Figure 4: The *series rule*.

Another special case is the *triangle rule*, shown in Figure 5.

**Reduction Rule 2b: Triangle rule**
Let $v$ be a vertex of degree 3 such that at least two of its neighbours are adjacent.
If $low \geq 3$, then
　add an edge between every pair of non-
　adjacent neighbours of $v$;
　remove $v$.

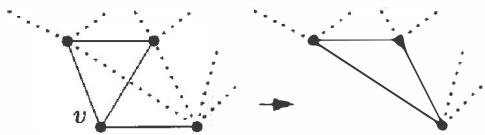

Figure 5: The *triangle rule*.

As the *series* and *triangle rules* are special cases of the *almost simplicial vertex rule*, both are safe.

If the *twig* and *islet rules* cannot be applied to a non-empty undirected graph, then the graph has treewidth at least two. We can then set *low* to $\max(low, 2)$. Note that from this observation we have that the *islet* and *twig rules* suffice for reducing any graph of treewidth one to the empty graph. The *islet*, *twig* and *series rules* suffice for reducing any graph of treewidth two to the empty graph. So, if $low \geq 2$ for a given non-empty graph and the *islet*, *twig* and *series rules* cannot be applied, then we know that the graph has treewidth at least three. We can then set *low* to $\max(low, 3)$.

As for treewidths one and two, there is a set of rules that suffice for reducing any graph of treewidth three to the empty graph. This set of rules was first identified by S. Arnborg and A. Proskurowski [7]. The *islet*, *twig*, *series* and *triangle rules* are among the set of six. The two other rules are of interest to us, not just because they provide for computing optimal triangulations for graphs of treewidth three, but also because they give new reduction rules.

**Lemma 3** *Let $G$ be an undirected graph and let $v, w$ be two vertices of degree three having the same set of neighbours. Let $G'$ be the graph that is obtained from $G$ by turning the set of neighbours of $v$ into a clique and then removing $v$ and $w$. Then,*

- $\tau(G') \leq \tau(G)$ and $\tau(G) \leq \max(3, \tau(G'))$;
- *the linear ordering $(v; (w; f))$, with $f$ a linear ordering of $G'$ of treewidth at most $\max(3, \tau(G'))$, has treewidth at most $\max(3, \tau(G'))$.*

**Proof.** Let $x$, $y$ and $z$ be the neighbours of $v$ and $w$. By contracting the edges $\{v, x\}$ and $\{w, y\}$ in $G$, we obtain $G'$. So, $G'$ is a minor of $G$. We find that $\tau(G') \leq \tau(G)$. Now, let $f$ be a linear ordering of $G'$ and let $H$ be the fill-in of $G'$ given $f$. If we add $v$ and $w$ with their (formerly) adjacent edges to $H$, then both are simplicial in the resulting graph. The treewidth of the ordering $(v; (w; f))$ of $G$, therefore, equals the maximum of 3 and the treewidth of $f$. The properties stated in the lemma now follow. □

From Lemma 3 we have safeness of the *buddy rule*, which is illustrated in Figure 6.

**Reduction Rule 3: Buddy rule**
Let $v, w$ be vertices of degree 3 having the same set of neighbours.
If $low \geq 3$, then
　add an edge between every pair of non-
　adjacent of neighbours of $v$;
　remove $v$;
　remove $w$.

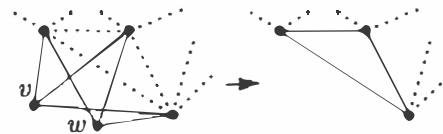

Figure 6: The *buddy rule*.

The *cube rule*, presented schematically in Figure 7, is slightly more complicated. The subgraph shown on the left is replaced by the one on the right; in addition, *low* is set to $\max(low, 3)$. Vertices $v$, $w$ and $x$ in the subgraph may be adjacent to other vertices in the rest of the graph; the four non-labeled vertices cannot have such 'outside' edges. Due to space limitations we do not provide a lemma from which safeness of the rule can be seen; the proof of safeness, however, is similar to the proofs given above.

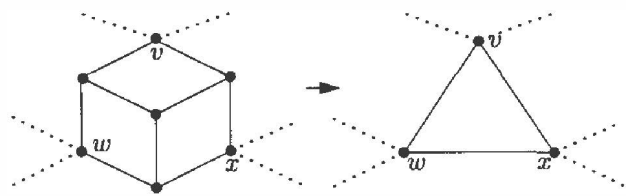

Figure 7: The *cube rule*.

The subgraph in the left hand side of the *cube rule* is not very likely to occur in the moralisation of a probabilistic network's digraph, but it is not impossible. The main reason for our interest in the rule is that it is one of the six rules that suffice for reducing graphs of treewidth three to the empty graph. So, if $low \geq 3$ for a given non-empty graph and the *islet, twig, series, triangle, buddy* and *cube rules* cannot be applied, then we know that the graph has treewidth at least four. We can then set $low$ to $\max(low, 4)$.

To conclude this section, Figure 8 shows a fragment of the well-known ALARM network, along with its moralisation. The figure depicts how application of our reduction rules serves to reduce the fragment to a single vertex. In fact, the moralised graph of the *entire* ALARM network is thus reduced to the empty graph, which indicates that our reduction rules provide for constructing an optimal triangulation.

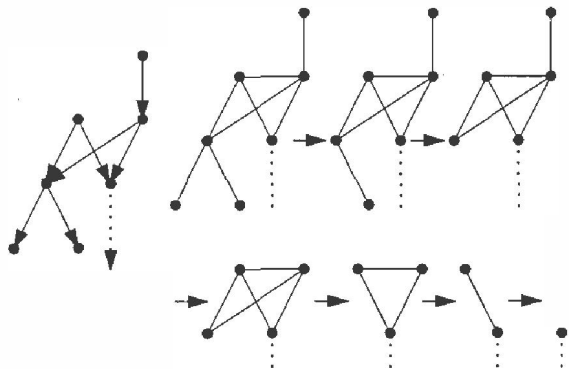

Figure 8: A fragment of the ALARM network and the reduction of its moralisation.

## 4 Computational method

The reduction rules described in the previous section are employed within a *computational method* that implements pre-processing of probabilistic networks for triangulation. We argued that application of our rules may reduce a network's moralised graph to the empty graph. The computational method complements this reduction by its reversal, thereby providing for the construction of a triangulation of minimal treewidth. For networks that cannot be triangulated optimally just by pre-processing, our rules are combined with an algorithm that serves to find an optimal or close to optimal triangulation of the reduced graph.

The computational method takes for its input the directed acyclic graph $G$ of a probabilistic network; it outputs a triangulation of the moralisation of $G$. The method uses a stack $S$ to hold the eliminated vertices in the order in which they were removed during the graph's reduction. Moreover, the value $low$ maintains a lower bound for the treewidth of the original moralised graph; it is initialised at 1. The method now amounts to the following sequence of steps:

1. The moralisation $M(G)$ of $G$ is computed and $G'$ is initialised at $M(G)$.

2. If a reduction rule can be applied to $G'$, it is executed, modifying $G'$ accordingly. Each vertex thus removed is pushed onto the stack $S$; if prescribed by the rule, the lower bound $low$ is updated. This step is repeated until the reduction rules are no longer applicable.

3. If no rule can be applied and $low < 4$, then $low$ is increased by 1. The reduction is continued at step 2.

4. Let $G'$ be the graph that results after execution of the previous steps. Using an exact or heuristic algorithm, $G'$ is triangulated.

5. Let $H$ be the triangulation that results from step 4. For $H$, a perfect elimination scheme $f$ is constructed.

6. Until $S$ is empty, the top element $v$ is popped from $S$ and $f$ is replaced by $(v; f)$.

7. Let $f'$ be the linear ordering resulting from the previous step. The fill-in of $M(G)$ given $f'$ is constructed.

The steps 1 through 3 describe the reduction of the graph of a probabilistic network. In step 4, the graph that results after reduction is triangulated. For this purpose, various different algorithms can be used. If the algorithm employed is *exact*, that is, if it yields a triangulation of minimal treewidth, then our method yields an optimal triangulation for the original moralised graph. For many real-life networks, the combination of our reduction rules with an exact algorithm results in an optimal triangulation in reasonable time. If after reduction a graph of considerable size remains for which an optimal triangulation cannot be feasibly computed, a *heuristic* triangulation algorithm can be used. The treewidth yielded for the original moralised graph then is not guaranteed to be optimal. As we will argue in the next section, however, these heuristic algorithms tend to result in better triangulations for the graphs that result from pre-processing than for the original graphs. If, after executing the steps 1 through 3, the reduced graph is empty, we can construct a triangulation of minimal treewidth for the moralised graph just by reversing the various reduction steps, and further triangulation is not necessary. This situation occurs, for example, if the original graph is already triangulated or has treewidth at most 3. The ALARM network gives another example: its moralised graph has treewidth four and is reduced to the empty graph.

In step 2 of our computational method, each of the reduction rules is investigated to establish whether or not it can be applied to the current (reduced) graph $G'$. As soon as an applicable rule is found, it is executed. When analysing the computational complexity of our method, it is readily seen that investigating applicability of the various reduction rules is the main bottleneck, as all other steps (except for the triangulation in step 4) take linear time [6].

Investigating applicability of the *islet, twig* and *series rules* takes a total amount of computation time that is linear in the number of vertices. To this end, we maintain for each vertex an integer that indicates its degree; we further maintain lists of the vertices of degree zero, one, and two. The *buddy*, *triangle* and *cube rules* can also be implemented to take overall linear time, for example using techniques from [9]. More straightforward implementations, however, will also be fast enough for moderately sized networks.

For the *simplicial vertex* and *almost simplicial vertex rules*, efficient implementation is less straightforward. To investigate whether or not a vertex is simplicial, we must verify that each pair of its neighbours are adjacent. For this purpose, we have to use a data structure that allows for quickly checking adjacency, such as a two-dimensional array. For a vertex of degree $d$, investigating simpliciality then takes $O(d^2)$ time. In a graph with $n$ vertices, we may have to check for simplicial vertices $O(n)$ times. Each such check costs $O(\sum_v d(v)^2) = O(ne)$ time, where $d(v)$ is the degree of vertex $v$ and $e$ denotes the number of edges in the graph. The total time spent on investigating applicability of the *simplicial vertex rule* is therefore $O(n^2 e)$. As the treewidth of the moralised graph of a real-life probabilistic network is typically bounded, we can refrain from checking simpliciality for vertices of large degree, giving a running time of $O(n^2)$ in practice. For the *almost simplicial vertex rule*, similar observations apply.

## 5 Experimental results

The computational method outlined in the previous section implements our method of pre-processing probabilistic networks for triangulation. We conducted some experiments with the method to study the effect of pre-processing. The results of these experiments are reported in this section.

The experiments were conducted on eight real-life probabilistic networks in the field of medicine: the WILSON network for the diagnosis of Wilson's liver disease; the well-known ALARM network for anaesthesia monitoring; the VSD network for the prognosis of ventricular septal defect in infants; the OESOPHAGUS network for the staging of oesophageal cancer and the extended OESOPHAGUS+ network for the prediction of response to treatment; the well-known MUNIN network for the interpretation of electromyographic findings; the ICEA network for the prediction of coverage by antibiotics of pathogens causing pneumonia; and the well-known PATHFINDER network for the diagnosis of lymphatic disease. The sizes of the digraphs of these networks and of their moralisations, expressed in terms of the number of vertices and the number of arcs and edges, respectively, are given in Table 1.

The effects of employing various different sets of pre-processing rules on the eight networks under study are summarised in Table 2. The sets employed are denoted:

$S$ = {*simplicial vertex*}
$PR$-1 = {*islet, twig*}

| network | before moralisation | | after moralisation | |
|---|---|---|---|---|
| | $\|V\|$ | $\|A\|$ | $\|V\|$ | $\|E\|$ |
| WILSON | 21 | 23 | 21 | 27 |
| ALARM | 37 | 44 | 37 | 62 |
| VSD | 38 | 51 | 38 | 61 |
| OESOPHAGUS | 42 | 57 | 42 | 68 |
| OESOPHAGUS+ | 67 | 117 | 67 | 194 |
| MUNIN | 1003 | 1244 | 1003 | 1662 |
| ICEA | 89 | 154 | 89 | 215 |
| PATHFINDER | 109 | 192 | 109 | 211 |

Table 1: The sizes of the digraphs of the various networks and of their moralisations.

$PR$-2 = $PR$-1 $\cup$ {*series*}
$PR$-3 = $PR$-2 $\cup$ {*triangle, buddy, cube*}
$PR$-4 = $S \cup PR$-3

With each of these sets of rules, the moralisations of the networks' graphs were reduced until the rules were no longer applicable. The table reports the sizes of the resulting reduced graphs. It reveals, for example, that the set of rules $PR$-3 suffices for reducing the moralised graphs of the WILSON and OESOPHAGUS networks to the empty graph; with the additional *simplicial vertex rule*, the moralised graphs of the ALARM and VSD networks are also reduced to the empty graph. These four networks are therefore triangulated optimally just by pre-processing. We would like to note that addition of the *almost simplicial vertex rule* to $PR$-4 did not result in any further reductions.

For the OESOPHAGUS+, MUNIN, ICEA and PATHFINDER networks, we further studied the effect of pre-processing on the treewidths yielded by various heuristic triangulation algorithms. Table 3 summarises the results obtained with *maximum cardinality search* (denoted MCS) [10] and with the *perfect-triangulation* and *minimal-triangulation* variants of *lexicographic breadth-first search* (denoted LEX_P and LEX_M, respectively) [11]. We ran the three heuristic algorithms on the original moralisations of the four networks and on the reduced graphs after employing the sets of rules $PR$-1 through $PR$-4. The table reveals that, for the MUNIN and ICEA networks, the heuristic algorithms tend to yield a smaller treewidth from the reduced graphs than from the original moralisations. The table in addition shows that the further reduced a graph, the less computation time is spent by the algorithms. As the time spent on pre-processing is negligible, these results indicate that pre-processing a probabilistic network is profitable, not just with respect to the quality of the triangulation yielded but also with respect to the computation time spent.

In our experiments we observed that the treewidths found by the heuristic algorithms depend to a large extent on the vertex with which the computation is started. To investigate the effect of the starting vertex, we ran the three heuristic algorithms a number of times, every time starting with a different vertex. Table 4 summarises the results, indicating per graph the minimum and maximum treewidth found and the average treewidth over all possible starting vertices. We would like to note that, using integer linear programming techniques on the most reduced graph, we found the exact treewidth of the PATHFINDER network to be 6.

| network | before pre-pro | | pre-pro with S | | pre-pro with PR-1 | | pre-pro with PR-2 | | pre-pro with PR-3 | | pre-pro with PR-4 | |
|---|---|---|---|---|---|---|---|---|---|---|---|---|
| | $|V|$ | $|E|$ | $|V|$ | $|E|$ | $|V|$ | $|E|$ | $|V|$ | $|E|$ | $|V|$ | $|E|$ | $|V|$ | $|E|$ |
| WILSON | 21 | 27 | 6 | 8 | 11 | 17 | 4 | 6 | 0 | 0 | 0 | 0 |
| ALARM | 37 | 62 | 7 | 11 | 30 | 55 | 13 | 28 | 5 | 10 | 0 | 0 |
| VSD | 38 | 61 | 16 | 21 | 22 | 45 | 12 | 28 | 6 | 14 | 0 | 0 |
| OESOPHAGUS | 42 | 68 | 5 | 8 | 22 | 48 | 12 | 28 | 0 | 0 | 0 | 0 |
| OESOPHAGUS+ | 67 | 194 | 28 | 125 | 46 | 173 | 31 | 144 | 28 | 135 | 26 | 121 |
| MUNIN | 1003 | 1662 | 449 | 826 | 819 | 1478 | 367 | 736 | 175 | 471 | 175 | 471 |
| ICEA | 89 | 215 | 64 | 176 | 85 | 211 | 66 | 181 | 59 | 170 | 59 | 170 |
| PATHFINDER | 109 | 211 | 14 | 49 | 68 | 170 | 37 | 112 | 17 | 63 | 14 | 49 |

Table 2: The effects of employing different sets of reduction rules.

| network | $|V|$ | $|E|$ | computed treewidth | | | CPU time spent | | |
|---|---|---|---|---|---|---|---|---|
| | | | MCS | LEX_P | LEX_M | MCS | LEX_P | LEX_M |
| OESOPHAGUS+_0 | 67 | 194 | 10 | 11 | 10 | 0.04 | 0.04 | 0.25 |
| OESOPHAGUS+_1 | 46 | 173 | 10 | 11 | 10 | 0.02 | 0.02 | 0.13 |
| OESOPHAGUS+_2 | 31 | 144 | 10 | 11 | 10 | 0.01 | 0.01 | 0.06 |
| OESOPHAGUS+_3 | 28 | 135 | 10 | 11 | 10 | 0.01 | 0.01 | 0.05 |
| OESOPHAGUS+_4 | 26 | 121 | 10 | 11 | 10 | 0.01 | 0.01 | 0.04 |
| MUNIN_0 | 1003 | 1662 | 10 | 16 | 16 | 28.60 | 35.78 | 330.90 |
| MUNIN_1 | 819 | 1478 | 10 | 16 | 16 | 17.99 | 23.12 | 216.29 |
| MUNIN_2 | 367 | 736 | 10 | 10 | 10 | 1.68 | 4.78 | 30.16 |
| MUNIN_3 | 175 | 471 | 9 | 8 | 8 | 0.29 | 0.37 | 3.29 |
| ICEA_0 | 89 | 215 | 15 | 14 | 13 | 0.12 | 0.14 | 0.72 |
| ICEA_1 | 85 | 211 | 15 | 14 | 13 | 0.11 | 0.13 | 0.65 |
| ICEA_2 | 66 | 181 | 16 | 15 | 13 | 0.09 | 0.09 | 0.42 |
| ICEA_3 | 59 | 170 | 15 | 14 | 13 | 0.08 | 0.08 | 0.33 |
| PATHFINDER_0 | 109 | 211 | 7 | 7 | 7 | 0.07 | 0.05 | 0.35 |
| PATHFINDER_1 | 68 | 170 | 7 | 7 | 7 | 0.02 | 0.02 | 0.15 |
| PATHFINDER_2 | 37 | 112 | 7 | 7 | 7 | 0.01 | 0.01 | 0.05 |
| PATHFINDER_3 | 17 | 63 | 7 | 7 | 7 | 0.00 | 0.00 | 0.01 |
| PATHFINDER_4 | 14 | 49 | 7 | 7 | 7 | 0.00 | 0.00 | 0.01 |

Table 3: The effect of pre-processing on the treewidths yielded by the three heuristic triangulation algorithms.

| network | $|V|$ | $|E|$ | MCS | | | LEX_P | | | LEX_M | | |
|---|---|---|---|---|---|---|---|---|---|---|---|
| | | | min | average | max | min | average | max | min | average | max |
| OESOPHAGUS+_0 | 67 | 194 | 10 | 12.1 | 13 | 11 | 12.8 | 16 | 10 | 12.0 | 14 |
| OESOPHAGUS+_1 | 46 | 173 | 10 | 12.2 | 15 | 11 | 12.8 | 16 | 10 | 12.1 | 14 |
| OESOPHAGUS+_2 | 31 | 144 | 10 | 11.9 | 15 | 11 | 12.7 | 16 | 10 | 11.8 | 13 |
| OESOPHAGUS+_3 | 28 | 135 | 10 | 12.1 | 15 | 11 | 12.6 | 16 | 10 | 11.6 | 13 |
| OESOPHAGUS+_4 | 26 | 121 | 10 | 11.8 | 15 | 11 | 12.6 | 16 | 10 | 11.6 | 13 |
| MUNIN_0 | 1003 | 1662 | 10 | 15.1 | 26 | 16 | 23.3 | 56 | 16 | 22.1 | 56 |
| MUNIN_1 | 819 | 1478 | 10 | 15.6 | 24 | 16 | 23.7 | 57 | 16 | 22.0 | 56 |
| MUNIN_2 | 367 | 736 | 10 | 14.8 | 25 | 10 | 22.1 | 50 | 10 | 20.7 | 50 |
| MUNIN_3 | 175 | 471 | 9 | 9.7 | 15 | 8 | 13.8 | 49 | 8 | 12.1 | 30 |
| ICEA_0 | 89 | 215 | 15 | 19.8 | 22 | 14 | 19.1 | 23 | 13 | 16.3 | 20 |
| ICEA_1 | 85 | 211 | 15 | 18.7 | 22 | 14 | 18.9 | 23 | 13 | 16.2 | 20 |
| ICEA_2 | 66 | 181 | 16 | 19.3 | 22 | 15 | 18.9 | 23 | 13 | 16.3 | 20 |
| ICEA_3 | 59 | 170 | 15 | 19.5 | 22 | 14 | 18.6 | 23 | 13 | 16.4 | 20 |
| PATHFINDER_0 | 109 | 211 | 7 | 7.2 | 8 | 7 | 8.0 | 9 | 7 | 7.6 | 8 |
| PATHFINDER_1 | 68 | 170 | 7 | 7.6 | 8 | 7 | 7.6 | 9 | 7 | 7.2 | 8 |
| PATHFINDER_2 | 37 | 112 | 7 | 7.5 | 8 | 7 | 7.6 | 9 | 7 | 7.2 | 8 |
| PATHFINDER_3 | 17 | 63 | 7 | 7.4 | 8 | 7 | 7.5 | 9 | 7 | 7.1 | 8 |
| PATHFINDER_4 | 14 | 49 | 7 | 7.6 | 8 | 7 | 7.4 | 9 | 7 | 7.1 | 8 |

Table 4: Some statistics for the three heuristic triangulation algorithms.

## 6 Conclusions and further research

When solving hard combinatorial problems, pre-processing is often profitable. Based upon this general observation, we designed a computational method that provides for pre-processing of probabilistic networks for triangulation. Our method exploits a set of rules for stepwise reducing the problem of finding a triangulation of minimum treewidth for a network's moralised graph to the same problem on a smaller graph. The smaller graph is triangulated, using an exact or heuristic algorithm, depending on the graph's size. From the triangulation of the smaller graph, a triangulation of the original graph is obtained by reversing the reduction steps. The reduction rules are guaranteed not to destroy optimality with respect to maximum clique size.

Experiments with our pre-processing method revealed that the graphs of some well-known real-life probabilistic networks can be triangulated optimally just by pre-processing. The experiments further showed that heuristic triangulation algorithms tend to yield better results for graphs that are reduced by pre-processing than for the original graphs. Moreover, the further reduced a graph, the less computation time is spent by the triangulation algorithms. From these observations, we conclude that pre-processing probabilistic networks for triangulation is profitable.

We are currently investigating other rules for pre-processing purposes. For example, D.P. Sanders designed a set of rules for reducing any graph of treewidth at most four to the empty graph [12]. Although this set is comprised of a large number of complex rules, it may give rise to new reduction rules that can be employed for pre-processing.

So far, we considered the use of rules for reducing the moralised graph of a probabilistic network. The use of separators constitutes another approach to pre-processing that we are currently investigating, building upon earlier work by K.G. Olesen and A.L. Madsen [13]. For example, if a network's moralised graph has a separator of size two, then the graph can be partitioned into smaller graphs that can be triangulated separately without losing optimality.

As there is a strong relationship between the running time of the junction-tree propagation algorithm and the treewidth of the triangulation used, most triangulation algorithms currently in use aim at finding a triangulation of minimal treewidth. However, if the variables in a probabilistic network have state spaces of diverging sizes, such a triangulation may not be optimal. A triangulation with minimal state space over all cliques then is likely to perform better. Some of our reduction rules are safe also with respect to minimum overall state space. Other rules, however, are safe only under additional constraints on their application. We plan to further investigate pre-processing for finding triangulations with minimum overall state space.

**Acknowledgements.** The research of the first author was partly supported by EC contract IST-1999-14186: Project ALCOM-FT (Algorithms and Complexity – Future Technologies). The research of the last two authors was partly supported by the Netherlands Computer Science Research Foundation with financial support from the Netherlands Organisation for Scientific Research.

## References

[1] S.L. Lauritzen and D.J. Spiegelhalter. Local computations with probabilities on graphical structures and their application to expert systems. *The Journal of the Royal Statistical Society. Series B*, vol. 50, pp. 157–224, 1988.

[2] F.V. Jensen, S.L. Lauritzen, and K.G. Olesen. Bayesian updating in causal probabilistic networks by local computations. *Computational Statistics Quarterly*, vol. 4, pp. 269–282, 1990.

[3] S. Arnborg, D.G. Corneil, and A. Proskurowski. Complexity of finding embeddings in a $k$-tree. *SIAM Journal on Algebraic and Discrete Methods*, vol. 8, pp. 277–284, 1987.

[4] K. Shoikhet and D. Geiger. A practical algorithm for finding optimal triangulations. *Proceedings of the National Conference on Artificial Intelligence (AAAI 97)*, pp. 185–190. Morgan Kaufmann, 1997.

[5] H.L. Bodlaender. A partial $k$-arboretum of graphs with bounded treewidth. *Theoretical Computer Science*, vol. 209, pp. 1–45, 1998.

[6] M.C. Golumbic. *Algorithmic Graph Theory and Perfect Graphs*. Academic Press, New York, 1980.

[7] S. Arnborg and A. Proskurowski. Characterization and recognition of partial 3-trees. *SIAM Journal on Algebraic and Discrete Methods*, vol. 7, pp. 305–314, 1986.

[8] A. Becker and D. Geiger. A sufficiently fast algorithm for finding close to optimal junction trees. *Proceedings of the Twelfth Conference on Uncertainty in Artificial Intelligence*, pp. 81–89. Morgan Kaufmann, 1996.

[9] S. Arnborg, B. Courcelle, A. Proskurowski, and D. Seese. An algebraic theory of graph reduction. *Journal of the ACM*, vol. 40, pp. 1134–1164, 1993.

[10] R.E. Tarjan and M. Yannakakis. Simple linear time algorithms to test chordality of graphs, test acyclicity of graphs, and selectively reduce acyclic hypergraphs. *SIAM Journal on Computing*, vol. 13, pp. 566–579, 1984.

[11] D.J. Rose, R.E. Tarjan, and G.S. Lueker. Algorithmic aspects of vertex elimination on graphs. *SIAM Journal on Computing*, vol. 5, pp. 266–283, 1976.

[12] D.P. Sanders. On linear recognition of tree-width at most four. *SIAM Journal on Discrete Mathematics*, vol. 9, pp. 101–117, 1996.

[13] K.G. Olesen and A.L. Madsen. Maximal prime subgraph decomposition of Bayesian networks. Technical report, Department of Computer Science, Aalborg University, Aalborg, Denmark, 1999.